\documentclass{llncs}

\usepackage{fancyhdr}
\usepackage{cite}
\usepackage{graphicx}
\usepackage{gensymb}
\usepackage{amsmath,amssymb,amsfonts}
\usepackage{multirow}
\usepackage{hyperref}

\DeclareMathOperator*{\argmin}{arg\,min}

\begin{document}
\title{Patch-Based Image Similarity for Intraoperative 2D/3D Pelvis Registration During Periacetabular Osteotomy}
%
%
\iftrue
\author{Robert B. Grupp\inst{1} \and Mehran Armand\inst{2,3} \and Russell H. Taylor\inst{1}}
\institute{Johns Hopkins University, Department of Computer Science, Baltimore, MD, USA\\
           \email{grupp@jhu.edu} \and
           Johns Hopkins University, Department of Mechanical Engineering, Baltimore, MD, USA \and
           Johns Hopkins Applied Physics Laboratory, Laurel, MD, USA
}
\else
\author{***********************************************************************}
\institute{*****************************************************************************\\
               *****************************************************************************\\
               *****************************************************************************\\
               *****************************************************************************\\
               *****************************************************************************
}
\fi

\maketitle 
\thispagestyle{fancy}
\pagestyle{fancy}

\begin{abstract}
Periacetabular osteotomy is a challenging surgical procedure for treating developmental hip dysplasia, providing greater coverage of the femoral head via relocation of a patient's acetabulum.
Since fluoroscopic imaging is frequently used in the surgical workflow, computer-assisted X-Ray navigation of osteotomes and the relocated acetabular fragment should be feasible.
We use intensity-based 2D/3D registration to estimate the pelvis pose with respect to fluoroscopic images, recover relative poses of multiple views, and triangulate landmarks which may be used for navigation. 
Existing similarity metrics are unable to consistently account for the inherent mismatch between the preoperative intact pelvis, and the intraoperative reality of a fractured pelvis.
To mitigate the effect of this mismatch, we continuously estimate the relevance of each pixel to solving the registration and use these values as weightings in a patch-based similarity metric.
Limiting computation to randomly selected subsets of patches results in faster runtimes than existing patch-based methods.
A simulation study was conducted with random fragment shapes, relocations, and fluoroscopic views, and the proposed method achieved a $1.7$ mm mean triangulation error over all landmarks, compared to mean errors of $3$ mm and $2.8$ mm for the non-patched and image-intensity-variance-weighted patch similarity metrics, respectively.
\end{abstract}
\section{Introduction}
\begin{figure}
	\centering
	\includegraphics[width=0.8\textwidth]{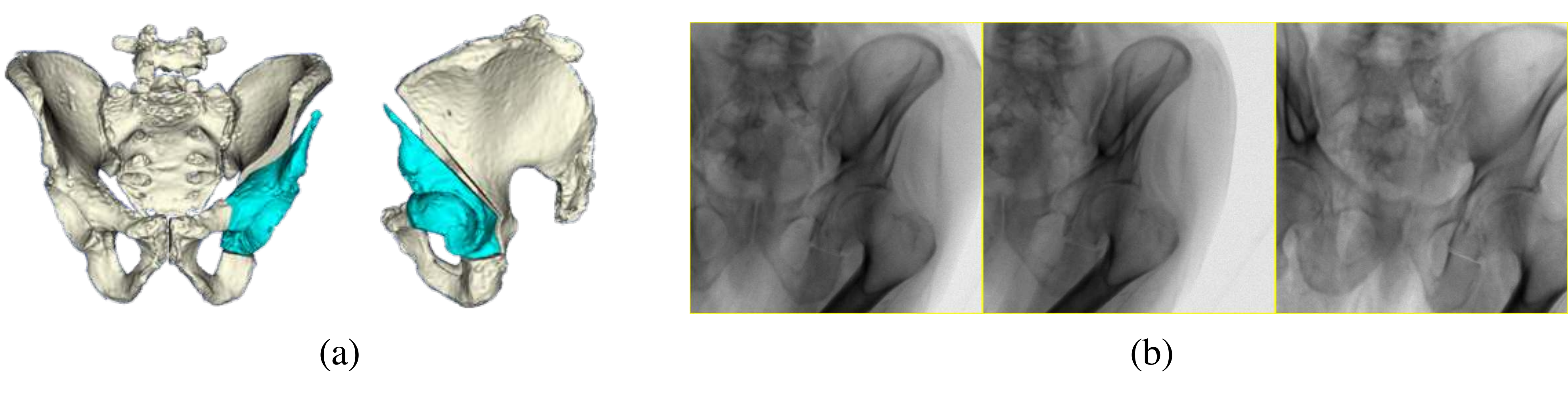}
	\caption{A simulated example of periacetabular osteotomies and a fragment reposition is shown in (a). The corresponding simulated fluoroscopic images are shown in (b).}
	\label{fig:frag_view_3d_xray_2d}
\end{figure}

Developmental dysplasia of the hip (DDH) is a condition with lower than normal coverage of the femoral head.
Patients with DDH frequently exhibit significant discomfort and are consequently less mobile.
Severe arthritis is a common long-term consequence of untreated DDH, therefore surgical treatment is expected during the lifetime of a patient \cite{murphy1995prognosis}.
The periacetabular osteotomy (PAO) is a surgical procedure designed to preserve the natural joint of young patients with DDH\cite{ganz1988new}.
In order to relocate the joint and increase femoral head coverage, the acetabulum must be freed from the remainder of the pelvis by performing osteotomies along the ilium, ischium, posterior column, and pubis.
Many clinicians use intraoperative fluoroscopy to manually navigate osteotomes while performing the cuts.
Even with fluoroscopic guidance, the ischial and posterior osteotomies introduce the risk of joint breakage due to their closeness to the acetabulum.
Furthermore, the fluoroscopic views are difficult to mentally interpret and accurate determination of femoral head coverage remains a challenge after its relocation \cite{troelsen2009surgical}.
A simulated set of PAO osteotomies with fragment movement, along with corresponding simulated fluoroscopic images, are shown in Fig. \ref{fig:frag_view_3d_xray_2d}.

Leveraging optical tracker navigation systems, several computer-assisted PAO approaches have been proposed to either track the osteotomes or estimate the pose of the acetabular fragment \cite{langlotz1998computer,murphy2015development,liu2016periacetabular}.
%
These systems require the attachment of at least one rigid body fiducial to a patient's bone and, in order to perform an accurate registration of the pelvis, require a tracked pointer tool to be swept across the surface of the relevant bone structures.
This requires a larger incision than typically used for PAO and eliminates the use of more modern, minimally invasive, approaches\cite{troelsen2008new}.
Taking into account these limitations and the prevalence of fluoroscopy use in PAO, we believe X-Ray navigation is a more prudent approach for computer-assisted navigation of osteotomes and bone fragment pose.
%

The pose of a patient's anatomy with respect to the fluoroscopy coordinate frame may be estimated using intensity-based 2D/3D, X-Ray/CT, registration \cite{markelj2012review}.
Using multiple views, 3D points with respect to the pelvis coordinate frame may be triangulated.
The motion of the fragment may be captured and reported by measuring the positions of landmarks prior to fragment relocation and afterwards.
Navigation of the osteotome with respect to the pelvis is feasible by estimating the locations of an osteotome's landmarks.
Most C-Arm models do not report the relative pose information of each view, therefore a fiducial object is typically used to establish a common coordinate frame and recover the multi-view geometry.
By using the pelvis as a fiducial, we avoid the introduction of new objects into the surgical workflow and fluoroscopic field of view.

Many methods exist to accurately register an intact pelvis with a single X-Ray view \cite{markelj2012review}, however PAO requires registration of the fractured pelvis with a relocated acetabulum.
Poor triangulation performance may result from irregular mis-registrations across views, since a fractured pelvis for PAO yields an intraoperative reality that is inconsistent with the preoperative model.
For example, registration of a particular view may be drawn to the pelvis fragment with iliac crest, while registration of another view may be drawn to the acetabulum.
In \cite{gong2011multiple} and \cite{joskowicz1998fracas}, 2D/3D registration of fractured bone fragments was proposed, both requiring preoperative CT of the bone fragments.
However, for PAO the fracture is created intraoperatively when 3D imaging is generally not available.
Manual masking of the model discrepencies in 2D is time consuming and will delay the surgical workflow.
By dividing the similarity computation across patches, and weighting each patch proportionally to the variance of image intensities, \cite{knaan2003effective} demonstrated registrations robust to the presence of metallic objects.
Since intensities corresponding to the relocated acetabular region have significant variance, this weighting is not effective for PAO pelvis registration.

In this paper, we use a preoperative weighting of 3D anatomical regions representing each region's expected contribution to an accurate registration of the fractured pelvis.
Using the current estimate of the pelvis pose, this weighting is projected into 2D at each optimization iteration and, after some additional processing, applied as weights for a patched similarity metric.
To the best of our knowledge, iterative adjustment of patch weightings has not been done in this way for 2D/3D registration.
By treating the patch weightings as a distribution over the most useful pixels during registration, computation in early optimization iterations may be restricted to small random subsets of patches, resulting in reduced runtimes.
These methods were evaluated with a simulation study accounting for various fragment shapes and movements.
With respect to rotation and translation registration errors and landmark triangulation error, the methods using iteratively adjusted weights outperformed existing similarity metrics.
\section{Methods}
\subsection{2D/3D Registration Overview}
The primary objective of X-Ray/CT, single-view, single-object, rigid registration is to compute the rigid transformation between the coordinate frame of a preoperative model and the coordinate frame of the X-Ray imager.
In this paper, we use an intensity-based registration approach, formulated as the optimization problem in \eqref{eq:regi_opt_prob}.
$\mathcal{S}$ represents a similarity metric between 2D images: the intraoperative fluoroscopic image, $I_X$, and a digitally reconstructed radiograph (DRR).
DRRs are created via the projection operator, $\mathcal{P}$, which uses a 3D volume of attenuations, $I_{CT}$, and the volume's pose with respect to the imaging coordinate frame, $\theta$.
$\mathcal{R}$ applies a regularization in order to penalize less plausible poses.
\begin{equation} \label{eq:regi_opt_prob}
	\argmin_{\theta \in SE(3)}  \mathcal{S}\left(I_X, \mathcal{P} \left( I_{CT}; \theta \right) \right) + \mathcal{R} \left( \theta \right)
\end{equation}
In this paper, we follow a multi-resolution approach, solving \eqref{eq:regi_opt_prob} at a low 2D resolution and using that solution as the initialization for \eqref{eq:regi_opt_prob} at a higher resolution.
The initialization to the low resolution level is determined by a 2D/3D paired-landmark registration.
We follow the approach of \cite{otake2012intraoperative} and use the CMA-ES optimizer at the lower resolution.
At the second resolution, the BOBYQA optimization algorithm is used and regularization is replaced with box constraints.
The object pose is parameterized by the $\mathfrak{se}(3)$ Lie algebra with $SE(3)$ reference point at the previous estimate of the object's pose with respect to the perspective projection coordinate frame and a center of rotation about the volume center.

In this work we only consider square shaped patches, defined by the center row, $c_r$, center column, $c_c$, and radius, $r$.
A computation over an entire image is equivalent to computation on a single image patch with size equal to the entire image extent.
The Normalized cross-correlation (NCC) similarity metric over a patch is defined in \eqref{eq:ncc}.
\begin{equation} \label{eq:ncc}
	\mathcal{S}_{NCC} \left( I_1, I_2 ; c_r, c_c, r \right) = \sum_{i = c_r - r}^{c_r + r} \sum_{j = c_c - r}^{c_c + r} 
																						\frac{\left( I_1\left( i,j \right) - \mu_{I_1} \right) \left( I_2\left( i,j \right) - \mu_{I_2} \right)}
																							   {\sigma_{I_1} \sigma_{I_2} \left( 2r + 1 \right)^2}
\end{equation}
Within the patch, the means of image intensities are denoted by $\mu_{I_1}$ and $\mu_{I_2}$; $\sigma_{I_1}$ and $\sigma_{I_2}$ denote the corresponding within patch standard deviations.
NCC assumes a linear relationship between the intensity values of each image, which is not satisfied by paired intraoperative fluoroscopy and DRRs derived from a CT with a single effective energy.
Computing NCC on the Sobel X and Y derivatives of the 2D images attempts to overcome this limitation and is defined in \eqref{eq:gncc}.
\begin{equation} \label{eq:gncc}
	\begin{aligned}
	\mathcal{S}_{GNCC} \left( I_1, I_2 ; c_r, c_c, r \right) & =  &  \mathcal{S}_{NCC} \left( \nabla_X I_1 , \nabla_X I_2 ; c_r, c_c, r \right) + \\
																	  				&	& \quad	\mathcal{S}_{NCC} \left( \nabla_Y I_1 , \nabla_Y I_2 ; c_r, c_c, r \right) 
	\end{aligned}
\end{equation}
Computing \eqref{eq:gncc} with a single patch, of size equal to the 2D image extent, shall be referred to as Grad-NCC.
Calculating \eqref{eq:gncc} over a set of patches distributed over the image, and combining the values in a weighted sum is shown in \eqref{eq:patchgncc}.
\begin{equation} \label{eq:patchgncc}
	\mathcal{S}_{PGNCC} \left( I_1, I_2 ; P\left( r \right), w \right) = \sum_{( k, l) \in P(r)} w\left(k,l \right) \mathcal{S}_{GNCC} \left( I_1, I_2 ; k, l, r \right)
\end{equation}
The set of all patch centers available within an image with patch radius, $r$, is defined as $P_{\text{complete}}(r)$.
%
%
The similarity metric using $P_{\text{complete}}(r)$ with a constant weighting shall be referred to as P-Grad-NCC.
Non-uniform patch weightings are used to emphasize that specific pixels should have more influence over the registration process.
As in \cite{knaan2003effective}, the variances of image intensities within patches may be used as a weighting; this method will be referred to as P-Grad-NCC-Var.
\subsection{Iterative Calculation of Patch Weights}
%
An ideal weighting will have largest values at 2D locations that are both feature-rich and consistent with the preoperative model.
Weights at locations expected to confound the registration will be assigned lower values.
To help achieve this, we rely on a preoperative 3D labeling which divides the pelvis into regions that are expected to produce useful and model-consistent features when forward projected.
A 3D weight image is computed using a manually specified lookup table defined over 3D labels.
Regions about the iliac crest, pubis ramus, sacrum-ilium junction, and vertebrae are assigned large weights, while areas corresponding to the ilium wing and soft-tissue are given smaller weights.
Very low weights are assigned to regions which are expected to change intraoperatively, such as the femur or any potential location on the acetabular fragment.

Several projection operations are performed using the current estimate of the pelvis pose.
A mask, $M$, of 2D pixel locations where it is likely for a mismatch with our preoperative model to occur, is computed by casting rays and checking for collision with possible fragment or femur regions in the preoperative plan.
A 2D boundary edge map of the pelvis is derived by checking for rays which intersect the intact pelvis' surface, and also have an adjacent ray not intersecting the surface.
Edges overlapping with expected mismatch locations in $M$ are pruned.
Next, the edge map is dilated and pruned once more.
An initial 2D weighting is produced through a maximum intensity projection of the 3D weight image.
Every 2D weight value corresponding to an edge pixel is scaled by $10$.
This allows edge features consistent with the model to dominate the registration.
Weight values at locations overlapping in $M$ are scaled by $0.1$.
This effectively serves as an automatic masking of regions which are believed to be inconsistent with our preoperative pelvis model.
The set of 2D weights is normalized to sum to 1.

\begin{figure}
	\centering
	\includegraphics[width=0.7\textwidth]{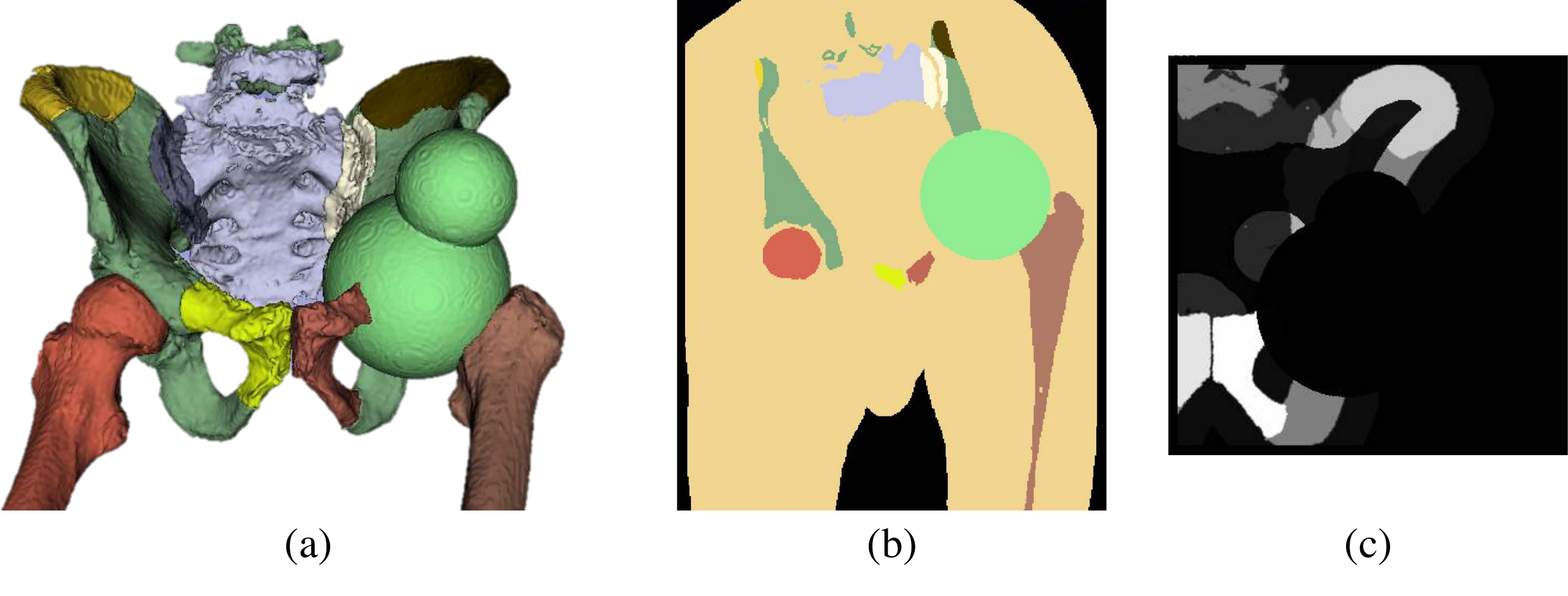}
	\caption{The surface rendering of a preoperative set of 3D labels used for iterative weight computation is shown in (a); (b) depicts a corresponding coronal slice.
				 The light green spheres encompass possible acetabular fragment and femoral head relocations, and are most likely to project to 2D pixels inconsistent with the preoperative model.
				 The iliac crest, left and right pubis rami, vertebrae, ilium wing, soft-tissue, etc. are all assigned different labels, allowing for a diverse assignment of 3D weightings.
				 An example of the 2D weights is shown in (c).
				 In areas expected to represent a relocated acetabular fragment, the femur, soft-tissue, or air, the weightings are very low.
				 Areas expected to contain the pubis ramus and iliac crest are weighted the largest, since we believe those features will be most helpful.
				 Other areas, such as vertebrae are given moderate weights, since they are expected to help with registration, but not be as helpful as the iliac and pubis regions.}
	\label{fig:labels_3D_proj_2D}
\end{figure}

The method employing this strategy is referred to as P-Grad-NCC-Pr.
The 3D preoperative labels used throughout this paper, along with a corresponding 2D weighting, is shown in Fig. \ref{fig:labels_3D_proj_2D}.
\subsection{Randomly Selecting a Subset of Patches}
\begin{figure}
	\centering
	\includegraphics[width=0.9\textwidth]{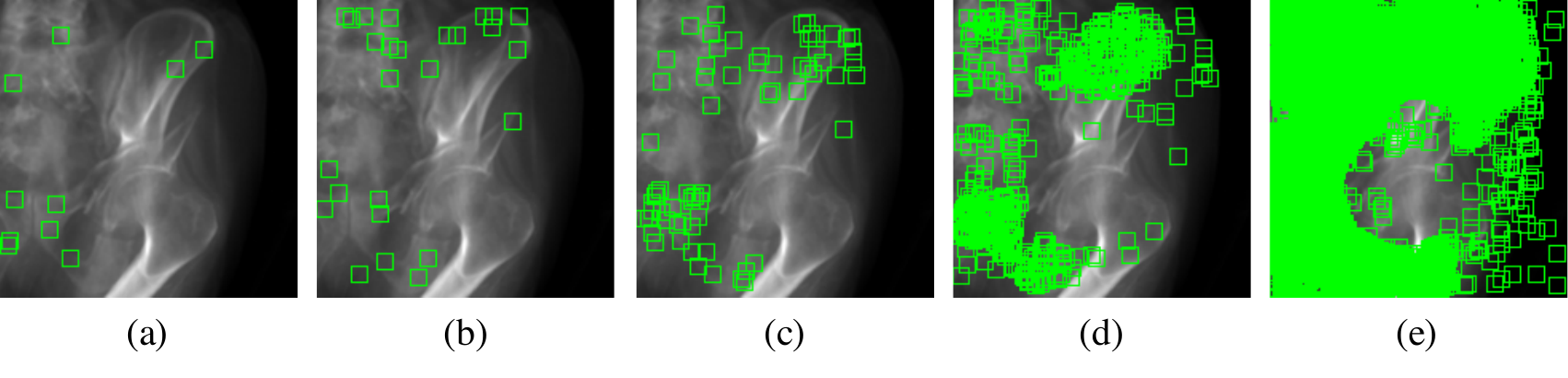}
	\caption{Randomly sampled patches used during a registration.
				(a) iteration 1; 10 patches.
				(b) iteration 3; 26 patches.
				(c) iteration 5; 68 patches.
				(d) iteration 9; 466 patches.
				(e) iteration 17; 21,892 patches.
				All patches were used after iteration 17.
				In (a) - (d), the patches are concentrated in areas consistent with the preoperative model, and which also have strong edge features and high contrast.}
	\label{fig:patches_overlay}
\end{figure}

We may treat the complete set of weightings as a categorical distribution over the available patches in an image.
A subset of patches, $P' (r) \subseteq P_{\text{complete}}(r)$ may be sampled using this distribution.
An updated weighting is obtained by re-normalizing the subset of original weights corresponding to the patches in $P'(r)$.
%
%
Computation is restricted to patches that are perceived to contain the most useful information for registration, by iteratively calculating weightings and then sampling random patches.
In order to achieve convergence, after each optimization iteration the number of patches is grown by a factor equal to the golden ratio: $(1 + \sqrt{5}) / 2$.
Once the number of random patches exceeds the maximum number of patches in the image, the metric reverts to using all patches.
Random patch sampling is only incorporated at the lower resolution level; the full set of patches is used at the second resolution level.
This method using randomly selected patches is referred to as P-Grad-NCC-Pr-R. 
An example of randomly selected patches during a registration is shown in Fig. \ref{fig:patches_overlay}.
\subsection{Simulated Data}
Simulated data is derived from pre and postoperative CT scans of a cadaveric specimen (male, 88 years), for which a PAO was performed by an experienced clinician.
Initial segmentations of the preoperative pelvis and femurs were obtained through an automated method \cite{krvcah2011fully}, and refined manually.
A rigid registration was performed to map the postoperative CT to the preoperative CT.
Points along each of the osteotomies in the postoperative CT were manually digitized and transformed into the preoperative coordinate frame.
Planes were fit to the transformed osteotomy points to obtain a baseline set of osteotomies.
The segmentation of the acetabular fragment is determined by the set of pelvis labels contained within the convex hull defined by the cutting planes.
Various fragment shapes were created by randomly rotating each cutting plane normal and translating by a random amount in the updated normal direction.
Collision detection against other bones was conducted to ensure randomly sampled movements of the fragment and femur were valid.
Soft-tissue is incorporated into the fluoroscopic image simulation by warping fragment and femur voxels within the volume and overwriting any overlapping soft-tissue voxels.
Random intensities in the HU range of muscle are used to fill any ``holes'' left by relocating the acetabulum and femur.
%
Fluoroscopic images were simulated similar to the procedure described in \cite{markelj2010standardized}.
Fig. \ref{fig:frag_view_3d_xray_2d} shows a relocated simulated fragment, and the corresponding set of 2D fluoroscopic images.
\subsection{Evaluation Metrics}
Registration rotation and translation errors are reported in the perspective projection coordinate frame with center of rotation located at the true location of the volume centroid.
The anatomical landmarks used for triangulation evaluation were the relocated femoral head (FH), the anterior superior iliac spine (ASIS), anterior inferior iliac spine (AIIS), greater sciatic notch (GSN), inferior obturator foramen (IOF), and the superior pubis symphysis (SPS).
These landmarks are useful as they correspond to the relocated fragment and measurement of possible BB locations, or are in close proximity to possible osteotomies and the measurement of osteotome positions.
For each fragment movement, the relative poses of three fluoroscopic views were estimated using pelvis registration transformations. The previous imaging world frame was replaced with the pelvis frame and each landmark position was triangulated.
%
%
\section{Experiments and Results}
\subsection{Simulation Study Parameters}
CT scans were acquired using a Toshiba Aquilion One with both $0.5$ mm slice spacing and thickness, and resampled to 1 mm isotropic spacings.
Using the left side of the specimen, 15 random fragments were sampled, and 20 random movements were sampled for each fragment.
Three fluoroscopy images were simulated from soft-tissue volumes created for each fragment movement.
The first view was initialized as an anterior-posterior view, followed by a random perturbation of the pelvis pose.
To obtain the second and third views, random orbital rotations in opposite directions were applied to the first view, followed by a small rigid perturbation.
This resulted in a total of 900 simulated fluoroscopy images.
%

Five random registration initializations were created for each fluoroscopic image by simulating a point picking process, followed by a landmark-based registration.
Human error was simulated by adding random noise to each 3D landmark and to each landmark visible in the 2D image.
Each initialization was used to run a registration for the following similarity metrics: Grad-NCC, P-Grad-NCC, P-Grad-NCC-Var, P-Grad-NCC-Pr, P-Grad-NCC-Pr-R.
A total of $4500$ total registrations per similarity metric were completed.

Simulated fluoroscopic images were $1536 \times 1536$ pixels, with $0.194$ mm/pixel isotropic spacing.
A source to detector distance of $1020$ mm and principal point at the center of the detector were used.
%
%

For CMA-ES, a population size of $100$ was used for all registrations, across all similarity metrics.
Downsampling of $8\times$ was done in each 2D dimension for the CMA-ES stage and $4\times$ for the BOBYQA stage.
Patches of size $11\times11$ pixels were used at the lower resolution level, and patches of $19\times19$ pixels were used at the higher resolution level.
The initial number of random patches used at the lower resolution level was 10.
%
%
Computation of DRRs and the Grad-NCC similarity metric were performed on the GPU.
The remainder of the similarity metrics were parallelized CPU implementations.
All registration trials were computed with dual Intel Xeon E5-2690 v2 CPUs and a single NVIDIA GeForce GTX TITAN Black GPU.
\subsection{Simulation Study Results}
%
%
Fourteen registration trials were discarded, corresponding to initialization offsets greater than $20\degree$ or $100$ mm.
Single-tailed Mann-Whitney U-Tests were performed to compare the errors of P-Grad-NCC-Pr and the remaining methods.
Acceptance of the alternative hypothesis indicated that the errors of P-Grad-NCC-Pr were drawn from a distribution with smaller median than the errors of the other method.
A p-value threshold of $0.005$ was used in each test.

The rotation and translation components of the initialization and registration errors are shown in Table \ref{tab:regi_errors}.
Each similarity metric performed well with respect to rotation, all with mean rotation error angles less than $1\degree$, however the patched similarity metrics with forward projected weights had the smallest mean rotation errors.
With respect to the total rotation angle error, there was no statistical difference between P-Grad-NCC, P-Grad-NCC-Pr, and P-Grad-NCC-Pr-R, however significantly larger errors were indicated for Grad-NCC and P-Grad-NCC-Var.
%
%
%
%
%
%
%
%
%
Most translation error was found in the depth direction (Z).
The patched similarity metrics achieved the best performance with respect to mean translation errors.
No statistical differences were indicated for the total translation errors of P-Grad-NCC, P-Grad-NCC-Pr, and P-Grad-NCC-Pr-R.
Grad-NCC and P-Grad-NCC-Var both had statistically larger total translation errors than P-Grad-NCC-Pr.
\begin{table}
\caption{Rotation/translation offsets from ground truth.
Rotation units are degrees and translation units are mm.
Statistically significant results are indicated with $*$.}
\label{tab:regi_errors}
\resizebox{\textwidth}{!}{%
\begin{tabular}{| c | c | r | r | r | r | r | r | }
\hline
 \multicolumn{2}{|c|}{Component}  & Initialization                      & Grad-NCC                          &     P-Grad-NCC                   &    P-Grad-NCC-Var              &   P-Grad-NCC-Pr     &  P-Grad-NCC-Pr-R  \\
\hline
\multirow{4}{*}{\rotatebox{90}{Rot.}}
& Total             & $2.0 \pm 1.2*$                   &  $0.6 \pm 0.5*$                  &  $0.4 \pm 0.8\enspace$      &    $0.7 \pm 0.9*$                  &    $0.4 \pm 0.7$        &   $0.4 \pm 0.7$   \\
& X                  &  $1.1 \pm 1.0*$                  &  $0.3 \pm 0.5*$                  &  $0.3 \pm 0.6*$                    &    $0.4 \pm 0.7*$                 &    $0.3 \pm 0.6$        &   $0.3 \pm 0.6$   \\
& Y                  & $1.1 \pm 0.9\enspace$      &  $0.4 \pm 0.4*$                  &  $0.2 \pm 0.5*$                   &    $0.3 \pm 0.5*$                  &    $0.2 \pm 0.4$        &   $0.2 \pm 0.4$   \\
& Z                  & $0.9 \pm 0.8\enspace$      &  $0.2 \pm 0.2\enspace$     &  $0.1 \pm 0.3\enspace$      &    $0.2 \pm 0.4\enspace$     &    $0.1 \pm 0.2$        &   $0.1 \pm 0.2$   \\
\hline
\multirow{4}{*}{\rotatebox{90}{Trans.}}
& Total             & $13.5 \pm 10.9*$ & $3.5 \pm 4.6*$                    &  $2.6 \pm 5.0\enspace$        &     $4.0 \pm 6.4*$                  &    $2.3 \pm 4.6$        &   $2.3 \pm 4.6$   \\
& X                  & $1.0 \pm 1.2*$     & $0.4 \pm 0.5\enspace$       &  $0.3 \pm 0.7\enspace$        &     $0.5 \pm 0.9\enspace$     &    $0.2 \pm 0.6$        &   $0.2 \pm 0.6$   \\
& Y                  & $1.1 \pm 1.0*$     & $0.6 \pm 0.7\enspace$       &  $0.4 \pm 0.7*$                     &    $0.7 \pm 0.9*$\                  &    $0.4 \pm 0.6$        &   $0.4 \pm 0.6$   \\
& Z                  & $13.3 \pm 10.9*$ & $3.3 \pm 4.6\enspace$       &  $2.5 \pm 5.0\enspace$        &    $3.8 \pm 6.4\enspace$      &    $2.2 \pm 4.6$        &   $2.2 \pm 4.6$    \\
\hline
\end{tabular}}
\end{table}
\begin{table}
\caption{Landmark triangulation errors from ground truth for initialization and each similarity metric. Units are mm.
Statistically significant results are indicated with $*$.}
\label{tab:trian_errors}
\resizebox{\textwidth}{!}{%
\begin{tabular}{| c | r | r | r | r | r | r | }
\hline
 Landmark   & Initialization         &           Grad-NCC            &     P-Grad-NCC                    &    P-Grad-NCC-Var    &   P-Grad-NCC-Pr     &  P-Grad-NCC-Pr-R  \\
\hline
FH              &  $6.1 \pm 6.3*$      &  $1.9 \pm 5.5*$             &   $1.6 \pm 5.3*$                   &   $2.4 \pm 5.8*$          &  $1.6 \pm 5.6$         &  $1.6 \pm 5.8$          \\
ASIS          &  $13.0 \pm 11.7*$  &   $4.7 \pm 8.5*$             &   $3.2 \pm 8.7*$                   &   $3.7 \pm 8.7*$          &  $2.9 \pm 8.8$         &   $2.9 \pm 8.8$        \\ 
AIIS            &  $9.5 \pm 9.2*$     &   $3.6 \pm 7.3*$             &   $2.5 \pm 7.4*$                   &   $3.1 \pm 7.5*$          &  $2.3 \pm 7.6$         &  $2.3 \pm 7.7$          \\ 
GSN           &  $5.5 \pm 4.1*$     &   $1.5 \pm 1.2*$             &   $0.9 \pm 1.7\enspace$      &   $1.4 \pm 1.9*$          &  $1.0 \pm 2.0$         &  $1.0 \pm 2.0$           \\ 
IOF             &  $4.1 \pm 3.1*$     &   $1.6 \pm 1.7*$             &    $1.1 \pm 2.1*$                  &   $2.3 \pm 3.4*$         &   $0.9 \pm 1.9$         &  $0.9 \pm 1.9$           \\
SPS            &  $4.0 \pm 3.0*$     &   $4.8 \pm 2.8*$             &    $2.0 \pm 1.8*$                  &   $3.8 \pm 3.2*$         &   $1.6 \pm 1.5$         &  $1.6 \pm 1.4$           \\ 
Combined   &  $7.0 \pm 7.7*$     &   $3.0 \pm 5.5*$             &    $1.9 \pm 5.4*$                  &   $2.8 \pm 5.7*$         &   $1.7 \pm 5.5$         &  $1.7 \pm 5.5$           \\
\hline
\end{tabular}}
\end{table}

Landmark triangulation errors are summarized in Table \ref{tab:trian_errors}.
Grouping all landmarks together, P-Grad-NCC-Pr and P-Grad-NCC-Pr-R had the smallest mean errors and were not significantly different.
Considering individual landmarks except the GSN, P-Grad-NCC-Pr and P-Grad-NCC-Pr-R had the smallest mean errors.
The only landmark for which a non-forward projected method did not have a significantly larger result was the GSN.
%
ASIS and AIIS errors were larger than errors of the remaining landmarks.
We believe this is due to inconsistent misalignments of the anterior iliac spine (AIS) across the views used for triangulation.
Compared to the rami of the ischium and pubis, the AIS is oriented parallel to the viewing directions, causing AIS image features to have less influence on image similarity than features associated with the ischium and pubis.
%
%
%
%
%
%
%
%
%

The mean registration runtimes, in seconds, were
$2.5 \pm 0.5$, $8.0 \pm 0.8$, $7.9 \pm 0.9$, $8.4 \pm 2.5$, $6.9 \pm 3.0$,
for Grad-NCC, P-Grad-NCC, P-Grad-NCC-Var, P-Grad-NCC-Pr, P-Grad-NCC-Pr-R, respectively.
Using random subsets of patches yields a speedup while not sacrificing performance.
%
%
%
%
%
%
%
%
%
%
%
%
%
%
%
%
%
%
%
%
%
%
\section{Discussion and Conclusion}
Accurate registration of the fractured pelvis during PAO is an essential component of an X-Ray navigation system for osteotomes and fragment relocations.
Through simulation, we have demonstrated the feasibility of a pelvis registration which is robust to the mismatch between the preoperative pelvis model and the intraoperative fractured pelvis.
Patch weightings are updated during each optimization iteration, resulting in significantly improved registration and triangulation performance compared with two existing methods.
Using random subsets of patches when iteratively updating weights was shown to have equivalent performance to using all patches and also have shorter runtimes.

%
%
%
%
We believe that a careful GPU implementation of P-Grad-NCC-Pr-R should have runtimes on par, or quicker than, the runtimes of Grad-NCC.
The most significant speedup could be obtained by limiting DRR computation to only pixels used by the similarity metric.
At each iteration, the CMA-ES optimization evaluates a large number of objective functions, each requiring a DRR.
A population size of 100 was used, resulting in $3,686,400$ pixels per iteration.
In contrast, a maximum of $121,000$ pixels are required when using ten $11 \times 11$ patches; a reduction of $97\%$ in the number of pixels.
We originally used a fixed number of random patches for P-Grad-NCC-Pr-R, however this resulted in poor convergence and excessive runtimes.
Analysis should be conducted to determine the optimal growth factor, and why a growth factor is necessary.
Preoperative annotation and planning is time consuming, however this process may be automated by registering the preoperative CT to a statistical model.
We plan to perform validation studies against fluoroscopy from cadavers which have undergone PAO.
\bibliography{refs} 
\end{document}